\documentclass[conference]{IEEEtran}
\IEEEoverridecommandlockouts
\usepackage{cite}
\usepackage{amsmath,amssymb,amsfonts}
\usepackage{algorithmic}
\usepackage{graphicx}
\usepackage{textcomp}
\usepackage{xcolor}
\usepackage{graphicx}
\usepackage{verbatim}
\usepackage{array}
\usepackage{tabularx}
\usepackage{makecell}
\usepackage{caption}
\usepackage{float}
\usepackage{amsmath}
\usepackage{booktabs}
\usepackage{threeparttable} 
\usepackage{color}
\usepackage{mathtools}

\usepackage{mwe} 
\usepackage{tabularx}
\usepackage{tikz}
\usepackage{pgfplots}
\usetikzlibrary{patterns}
\pgfplotsset{compat=1.18} 

\def\BibTeX{{\rm B\kern-.05em{\sc i\kern-.025em b}\kern-.08em
    T\kern-.1667em\lower.7ex\hbox{E}\kern-.125emX}}
\begin{document}

\title{Personalized Counterfactual Framework: Generating Potential Outcomes from Wearable Data}

\author{Ajan Subramanian$^{1*}$ and Amir M. Rahmani$^{1,2}$ 
\thanks{$^{1}$Dept. of Computer Science, University of California, Irvine. 
        $^{2}$School of Nursing, University of California, Irvine. 
        {\tt\small (*correspondence email: ajans1@uci.edu)}}%
}

\maketitle

\begin{abstract}
Wearable sensor data offer opportunities for personalized health monitoring, yet deriving actionable insights from their complex, longitudinal data streams is challenging. This paper introduces a framework to learn personalized counterfactual models from multivariate wearable data. This enables exploring ``what-if'' scenarios to understand potential individual-specific outcomes of lifestyle choices. Our approach first augments individual datasets with data from similar patients via multi-modal similarity analysis. We then use a temporal PC (Peter-Clark) algorithm adaptation to discover predictive relationships, modeling how variables at time $t-1$ influence physiological changes at time $t$. Gradient Boosting Machines are trained on these discovered relationships to quantify individual-specific effects. These models drive a counterfactual engine projecting physiological trajectories under hypothetical interventions (e.g., activity or sleep changes). We evaluate the framework via one-step-ahead predictive validation and by assessing the plausibility and impact of interventions. Evaluation showed reasonable predictive accuracy (e.g., mean heart rate MAE 4.71 bpm) and high counterfactual plausibility (median 0.9643). Crucially, these interventions highlighted significant inter-individual variability in response to hypothetical lifestyle changes, showing the framework's potential for personalized insights. This work provides a tool to explore personalized health dynamics and generate hypotheses on individual responses to lifestyle changes.
\end{abstract}

\section{Introduction}
Wearable technologies enable continuous, personalized monitoring of physiological and behavioral data, offering significant potential for proactive health management aligned with 4P medicine (predictive, preventive, personalized, participatory) \cite{kristoffersson2020systematic, haghi2021wearable, hood2012personal}. These devices track parameters like heart rate variability (HRV), sleep, readiness, and activity, providing rich data for early detection and individualized interventions \cite{darwish2011wearable, babu2024wearable}. However, the high-volume, multimodal, and longitudinal nature of this data presents analytical challenges. Generating actionable individual-level insights necessitates models that capture personalized health dynamics beyond simple descriptive statistics \cite{lago2024state}.

While existing research leverages wearable data for predictive tasks \cite{yang2023loneliness, yang2024integrating, jafarlou2023objective, subramanian2024graph}, many approaches identify associations without elucidating underlying mechanisms and may not adequately address inter-individual variability. Effective personalized healthcare requires moving beyond generalized models to understand individual-specific responses and the interplay between factors like sleep, exertion, and autonomic function \cite{raita2021big, sanchez2022causal}. Critically, while associative models can predict likely future states, they often fall short of exploring how an individual might uniquely respond to specific changes in their behavior or environment. Understanding this is vital for truly actionable and personalized guidance. 

To address this, we propose a framework for personalized counterfactual modeling. Counterfactuals, or estimations of potential outcomes under hypothetical scenarios, provide a powerful approach to move beyond observed correlations. They allow for probing individual-specific cause-and-effect relationships that are essential for tailoring interventions, even if those exact scenarios have not been previously experienced by that individual. Our approach enhances individual datasets through patient similarity analysis to improve model robustness given limited data per person. We then employ a temporal adaptation of a constraint-based algorithm to estimate plausible predictive structures from observational time series, focusing on how variables influence daily changes ($\Delta V$) in an individual's physiological state \cite{moraffah2021causal}. These personalized models form the basis for counterfactual generations, allowing us to ask targeted ``what-if?'' questions (e.g., ``What would this individual's activity score likely have been today \textit{if} they had experienced poor sleep last night?''). By generating hypothetical scenarios, we estimate the potential effects of interventions based on learned individual-specific dynamics.

\begin{figure*}[htbp]
    \centering
    \includegraphics[width=\linewidth]{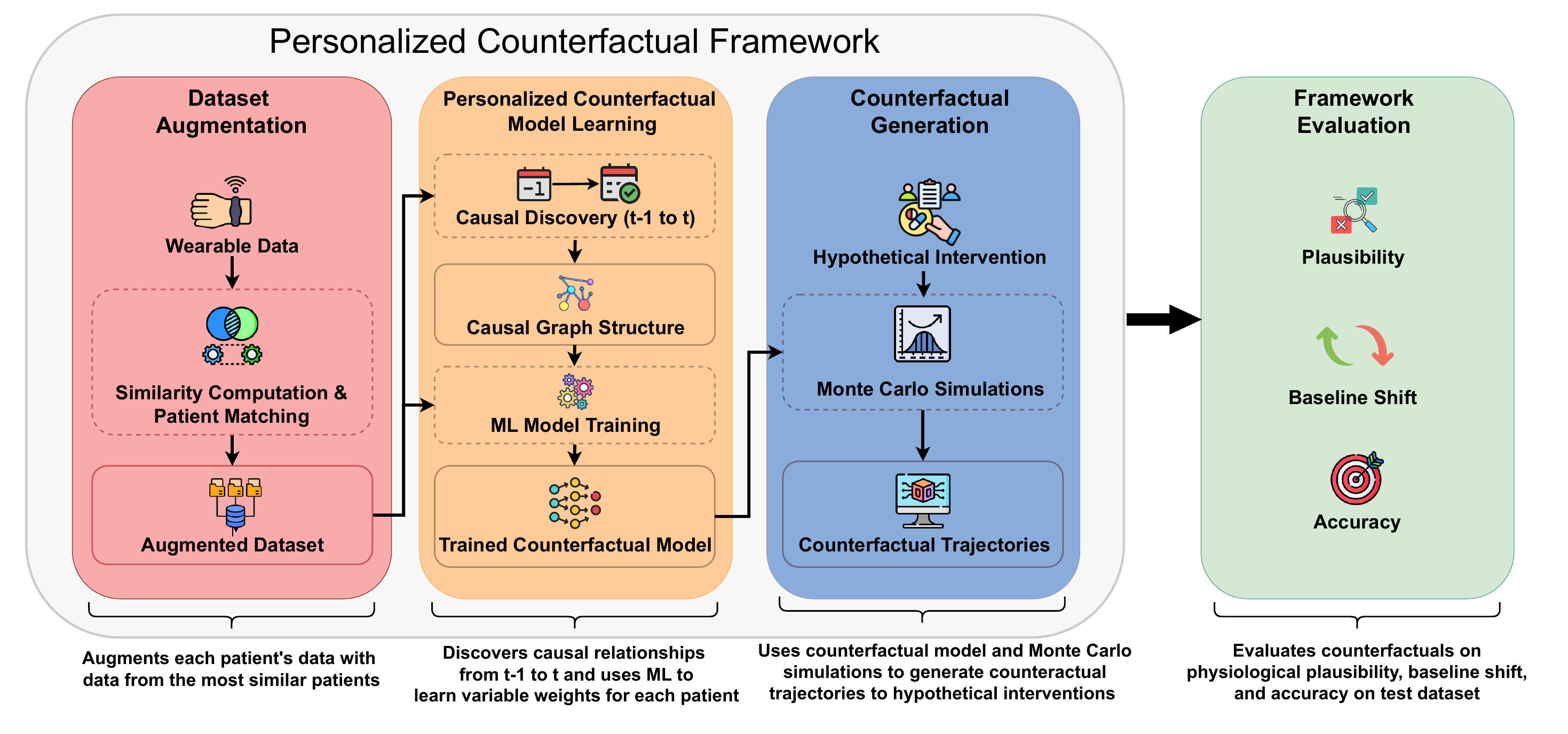}
    \caption{Overview of the personalized counterfactual framework.}
    \label{fig:framework_overview}
\end{figure*}

The key contributions of this paper are:
\begin{enumerate}
    \item A methodology for learning personalized counterfactual models from multivariate wearable data using a temporal PC algorithm and regression on variable changes ($\Delta V$).
    \item Integration of patient similarity analysis to augment individual datasets, improving model stability.
    \item A counterfactual generation framework leveraging these personalized models to estimate potential outcomes of hypothetical interventions.
    \item A comprehensive evaluation approach assessing both one-step-ahead predictive accuracy and the plausibility and impact of interventions.
\end{enumerate}

This framework prioritizes understanding individual-specific health responses and the comparative effects of interventions, rather than solely aiming for maximum predictive accuracy. It is intended as a tool to help generate new hypotheses about personalized health dynamics

\section{Method}

Our framework (Fig. \ref{fig:framework_overview}) involved four main stages: dataset augmentation, personalized counterfactual model learning, counterfactual generation, and framework evaluation.

\subsection{Dataset Augmentation} 
We analyzed data from 20 college students monitored for an average of 7.8 months during the 2020 COVID-19 lockdown as part of a remote health monitoring study~\cite{labbaf2024physiological}. The data collected included sleep metrics, activity, RMSSD, and readiness scores from Oura Rings and Samsung smartwatches, alongside demographics and self-reported sentiment/mood data. 

Preprocessing involved median imputation, standardization (numeric), one-hot encoding (categorical), and frequency summaries (sentiment/thematic annotations).
Patient similarity was computed using a weighted average of cosine similarities across feature types. Each patient's time series was then augmented with data from their three most similar peers. The patient’s own data received a higher fixed weight to emphasize individual characteristics, a process designed to stabilize personalized models, especially when individual historical data are limited. The output of this stage was an augmented dataset for each patient with 14 physiological features.

\subsection{Personalized Counterfactual Model Learning} 
A personalized counterfactual model (PCM) was constructed for each patient using their augmented dataset, modeling the change in physiological variables ($\Delta V_t = V_t - V_{t-1}$) as a function of lagged variables ($V_{t-1}$).

\textit{Causal Structure Discovery:}
We used a temporal adaptation of the PC constraint-based algorithm \cite{kalisch2007estimating} to identify plausible predictive relationships from $V_{t-1}$ to $\Delta V_t$. This iteratively tests for conditional independence between lagged variables ($V_{t-1, i}$) and target change ($\Delta V_{t,j}$) using weighted partial correlation while accounting for weights from the augmentation state. Edges in the causal structure graph, representing potential influences, were retained if their p-value fell below 0.05. This yielded sparse predictive graphs for each $\Delta V_{t,j}$ for each patient.

\begin{figure}[htbp]
\centering
\includegraphics[width=0.34\textwidth]{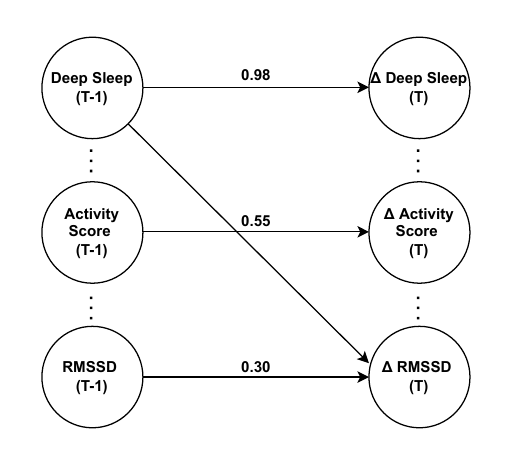} 
\caption{Illustrative predictive structure from a single patient's personalized model. Arrows depict learned influences of lagged variables $V_{t-1}$ on physiological changes $\Delta V_{t}$. The numerical values represent relative feature importances (derived from GBM) for key predictive links.}
\label{fig:causal_graph}
\end{figure}

\textit{Model Training:}
For each ($\Delta V_{t,j}$), a Gradient Boosting Machine (GBM) was trained to quantify the discovered relationships.  Fig. \ref{fig:causal_graph} provides an illustrative example from a single patient, showing such learned relationships and the relative feature importance of selected predictors influencing physiological changes.
The predictors for each GBM included the set of lagged variables ($V_{t-1,i}$) identified by the temporal PC algorithm for that specific $\Delta V_{t,j}$, plus a mandatory self-lag.  Prior to training, both predictor and target change variables were z-score normalized using statistics derived from the patient's augmented training data. GBM hyperparameters were optimized via grid search with cross-validation. The normalization statistics and the standard deviation of the standardized residuals from each GBM were stored.

\subsection{Counterfactual Generation} 
The learned models enabled the generation of potential outcomes to hypothetical interventions. Each patient's validated PCM, comprising their trained GBMs and associated parameters (normalization statistics and learned noise characteristics), drove an iterative process. To conduct counterfactual generation, an initial physiological state ($X_0$) was defined, typically from recent observed data. A hypothetical intervention (e.g., modifying `activity score') was applied to $X_0$.

The generation unfolded step-by-step over a defined time horizon. At each step $t$:
\begin{enumerate}
    \item The current state $X_t$ was standardized.
    \item The GBMs predicted the standardized change ($\Delta Z_{\text{pred}}$) based on the current standardized state.
    \item A stochastic element ($\epsilon_{t+1}$), sampled from the learned noise distribution, was added.
    \item The total change updated the standardized state to $Z_{t+1}$.
    \item $Z_{t+1}$ was unstandardized to $X_{t+1}^{\text{raw}}$, clipped to realistic physiological bounds, and $X_{t+1}$ became the input for the next iteration.
\end{enumerate}
To account for stochasticity from the noise term, Monte Carlo simulations were performed (multiple runs with different random noise samples), generating a distribution of potential future trajectories (the counterfactuals) for each scenario.

\subsection{Framework Evaluation} 
The overall framework was assessed through metrics evaluating model accuracy, as well as the plausibility and impact of the counterfactuals. Initially, to ensure the foundational PCMs accurately capture short-term dynamics, their \textbf{one-step-ahead predictive accuracy} was validated on hold-out data. This involved generating deterministic ( $ \epsilon=0 $) predictions and quantifying accuracy using Mean Absolute Error (MAE) and Root Mean Squared Error (RMSE), calculated between predicted and actual values. Subsequently, the \textbf{counterfactuals} themselves were evaluated. This involved assessing their \textbf{Plausibility}, defined as the fraction of generated time points where the 5th-95th percentile range of trajectories (from Monte Carlo runs) remains within predefined realistic physiological bounds. Furthermore, the \textbf{Baseline Shift} was measured; for a given intervention and outcome, this quantified the difference between the mean generated outcome at the horizon (averaged over Monte Carlo runs) and its baseline value.

\begin{figure*}[htbp]
\centering
\includegraphics[width=0.90\textwidth]{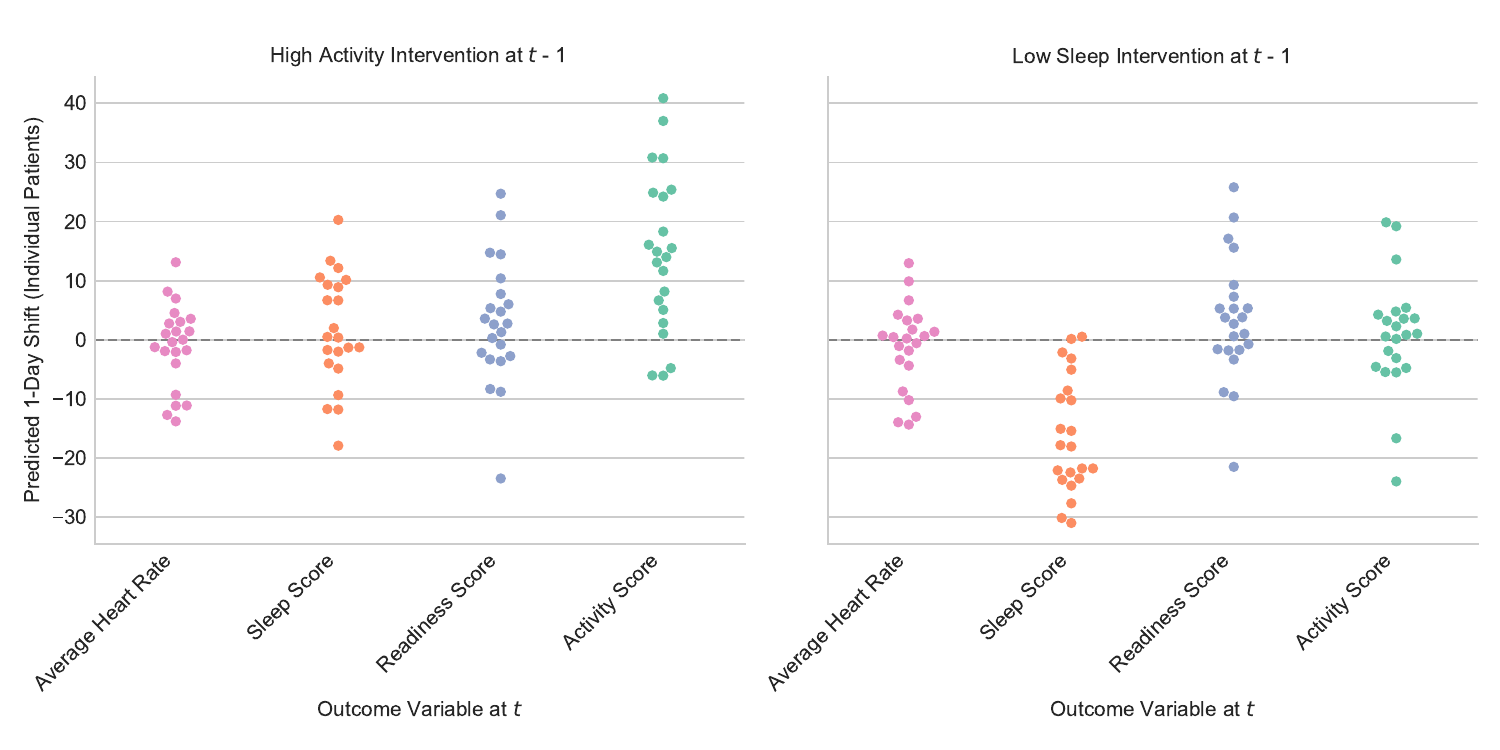} 
\caption{Heterogeneous patient responses to interventions. Each point represents a unique patient's predicted 1-day shift for the specified outcome variable under `High Activity' (left panel) and `Low Sleep' (right panel) scenarios. The swarm plots illustrate the personalized nature and variability of intervention effects on Average Heart Rate, Sleep Score, Readiness Score, and Activity Score. The dashed horizontal line indicates zero shift.}
\label{fig:personalized_effects}
\end{figure*}

\section{Results and Discussion}
\label{sec:results}
We evaluated the framework's one-step-ahead predictive accuracy and the plausibility and impact of its generated counterfactuals. The presented results are from models trained using GBMs on relationships discovered by the temporal PC algorithm, without explicit domain knowledge.

\subsection{Model Performance and Counterfactual Plausibility}
Personalized models demonstrated a reasonable ability to capture short-term physiological dynamics, with detailed predictive performance metrics provided in Table \ref{tab:predictive_performance}. For example, the mean MAE for average heart rate was 4.71 bpm, and for sleep score was 6.88 points.

\begin{table}[htbp]
\centering
\caption{One-Step-Ahead Predictive Performance (Mean (Std))}
\label{tab:predictive_performance}
\begin{threeparttable}
\begin{tabularx}{\columnwidth}{l >{\raggedleft\arraybackslash}X >{\raggedleft\arraybackslash}X}
\toprule
\textbf{Variable} & \textbf{MAE} & \textbf{RMSE} \\
\midrule
Average Heart Rate   & 4.71 (2.60) & 5.67 (3.36) \\
Readiness Score (0-100)     & 6.45 (2.46) & 7.69 (2.95) \\
Sleep Score (0-100)         & 6.88 (3.01) & 8.02 (3.24) \\
Activity Score (0-100)      & 7.07 (3.26) & 8.24 (3.76) \\
\bottomrule
\end{tabularx}
\end{threeparttable}
\end{table}

Generated physiological trajectories exhibited high counterfactual plausibility (median 0.9643, IQR: 0.9643 -- 1.0000 across scenarios), with trajectories largely remaining within realistic physiological bounds.

\subsection{Personalized Effects of Interventions}
A key strength of the framework is its ability to capture heterogeneous responses to interventions. Fig. \ref{fig:personalized_effects} visually underscores this patient-to-patient variability for the hypothetical `High Activity' and `Low Sleep' scenarios. Each point in the swarm plots represents an individual patient's predicted 1-day shift in key outcome variables.

For example, under the `High Activity' intervention, while the average `activity score' increased substantially (mean shift: +14.8 points), the individual responses varied widely. Some patients exhibited minimal change, whereas others showed large increases. Similarly, the `Low Sleep' intervention typically reduced `sleep score' (mean shift: -16.0 points), but the magnitude of this reduction differed considerably across individuals. This heterogeneity underscores the need for personalized models to anticipate individual reactions to lifestyle modifications, crucial for tailoring effective health recommendations. Collectively, the predictive accuracy, high plausibility, and the personalized intervention effects demonstrate the potential of our framework for generating actionable insights from wearable data. 

\section{Conclusion}
\label{sec:conclusion}
We presented a framework for creating personalized counterfactual models from wearable sensor data to explore individual responses to hypothetical interventions. Our approach combines patient similarity, temporal causal discovery for daily physiological changes, and predictive modeling with GBMs to generate plausible outcomes highlighting significant inter-individual variability in intervention effects. We evaluated our approach using accuracy, plausibility, and shifts from baseline data, underscoring its potential to move beyond generalized insights towards tailored health strategies. One limitation of our work is the assumption of stationarity in physiological variables over time. Future work will focus on validating these personalized counterfactuals against real-world interventions, extending time horizons, and exploring more complex dynamic interactions to further enhance personalized health forecasting and decision support from wearable technology. 

There is broader applicability to our framework beyond research. For instance, it can run as a lightweight cloud microservice behind consumer wearables, turning raw scores into personalized ``what-if'' nudges that encourage healthy habits. Moreover, embedded in remote patient-monitoring dashboards, it can help clinicians preview intervention impact. Ultimately, by translating complex sensor streams into actionable, patient-specific ``what-if'' insights, this framework lays the groundwork for a new generation of proactive, truly personalized digital health interventions.

\bibliographystyle{IEEEtran}
\bibliography{IEEEabrv, ref} 

\end{document}